%% file: neurips_2025.tex
\documentclass{article}

 \usepackage[preprint]{neurips_2025}
\usepackage{tikz}
\usetikzlibrary{positioning}
\usepackage[utf8]{inputenc} % allow utf-8 input
\usepackage[T1]{fontenc}    % use 8-bit T1 fonts
\usepackage{hyperref}       % hyperlinks
\usepackage{url}            % simple URL typesetting
\usepackage{booktabs}       % professional-quality tables
\usepackage{amsfonts}       % blackboard math symbols
\usepackage{nicefrac}       % compact symbols for 1/2, etc.
\usepackage{microtype}      % microtypography
\usepackage{xcolor}         % colors
\usepackage{amsmath}
\usepackage{array}
\usepackage{tabularx}
\title{The UNDO Flip-Flop: A Controlled Probe for Reversible Semantic State Management in State Space Models}

\author{%
  Hongxu Zhou \\
  Erasmus Mundus Language and Communication Technologies\\
  Saarland University\\
  Saarbrücken, Germany \\
  \texttt{hozh00003@stud.uni-saarland.de} \\
}

\begin{document}

\maketitle

\begin{abstract}
  State space models (SSMs) have been shown to possess the theoretical capacity to model both star-free sequential tasks and bounded hierarchical structures \cite{sarrofExpressiveCapacityState2024}. However, formal expressivity results do not guarantee that gradient-based optimisation will reliably discover the corresponding solutions. Existing benchmarks probe either monotonic state tracking, as in the standard Flip-Flop task, or structural nesting, as in the Dyck languages, but neither isolates reversible semantic state retrieval. We introduce the UNDO Flip-Flop task to fill this gap. By extending the standard Flip-Flop with an \texttt{[UNDO]}, the task requires a model to maintain an implicit bounded stack and recover historical states under non-monotonic update sequences. We evaluate one-layer and two-layer Mamba-2 under this framework. Both variants fail to acquire the provably expressible stack-based rollback mechanism, converging instead on a local toggle heuristic that inverts the current state rather than retrieving stored history. Under an adversarial retraction pressure test held within the training length distribution, the two-layer model collapses to 41.10\% accuracy being below random chance. The results confirm systematic rather than incidental failure. Causal ablation shows that the bottleneck lies in retrieval, not storage. These results draw a clear line between what an architecture can in principle represent and what gradient descent reliably learns, a distinction that theoretical expressivity analyses alone cannot capture. The code is avaiable here: \url{https://github.com/hongxuzhou/undo_flip_flop}
\end{abstract}

\input{sections/1intro}

\input{sections/2LR}

\input{sections/3Mthd}

\input{sections/4Experiments}

\input{sections/5Discussion}

\input{sections/6Conclude}

\bibliographystyle{plainnat}
\bibliography{undo_flip_flop}

\appendix
\input{sections/7Appendex}

\end{document}

%% file: sections/1intro.tex
% !TeX root = ../neurips_2025.tex
\section{Introduction}
The question of what sequence models can in principle compute is distinct from what they reliably learn under gradient descent. Recent theoretical work has placed both Transformers and linear SSMs within the $TC^0$ complexity class \citep{merrillIllusionStateStateSpace2025}, but within this shared ceiling, SSMs hold specific advantages: \citet{sarrofExpressiveCapacityState2024} prove that a two-layer SSM can model star-free sequential tasks with length-generalising solutions and track bounded hierarchical structure without simulating an explicit stack. These results establish representational capacity. They do not tell us whether a training algorithm will find those solutions.

Standard benchmarks probe this capacity only partially. The Flip-Flop task \citet{liuExposingAttentionGlitches2023} tests monotonic, destructive overwrite memory; Dyck languages test structural nesting. Neither isolates reversible semantic state retrieval: the ability to retract a prior update and restore a previously stored value. This operation appears wherever inference is not strictly cumulative: speech repair, multi-turn dialogue correction, and chain-of-thought reasoning that must abandon an intermediate conclusion and recover an earlier position all place the same demand on ordered historical access.

We introduce the UNDO Flip-Flop task to fill this gap, extending the standard Flip-Flop with a stack-based pop operator that requires the model to retrieve specific historical values under dynamic retractions. Evaluating Mamba-2 on this task, we find that neither a one-layer nor a two-layer configuration acquires the provably expressible stack mechanism. Both converge instead on a local toggle heuristic that inverts the current state rather than retrieving stored history — a shortcut that passes sparse in-distribution evaluation but collapses below random chance under adversarial retraction pressure. The primary contribution is an empirical distinction between architectural expressibility and what optimisation reliably extracts from it.

%% file: sections/2LR.tex
% !TeX root = ../neurips_2025.tex
\section{Related Work}
\label{sec:rl}
\subsection{Models and Formal Expressivity}
\label{sec:models-formal-expressivity}

The Transformer architecture achieves high scalability through self-attention \citep{vaswaniAttentionAllYou2017}. This design processes inputs simultaneously, which provides strong global association but fundamentally restricts strict sequential state tracking. Theoretical analyses show that log-precision Transformers are confined to the $TC^0$ complexity class and cannot reliably compute inherently sequential problems \citep{merrillParallelismTradeoffLimitations2023}. Self-attention mechanisms also struggle to evaluate periodic finite-state languages and strictly hierarchical structures \citep{hahnTheoreticalLimitationsSelfAttention2020}.

Linear State Space Models (SSMs) like Mamba address these limitations by reintroducing recurrent structures with selective state spaces, allowing dynamic information filtering \citep{guMambaLinearTimeSequence2024}. The subsequent Mamba-2 architecture improves hardware efficiency by linking SSMs and attention mechanisms through State Space Duality \citep{daoTransformersAreSSMs2024}. To achieve this, Mamba-2 simplifies its internal state matrix to a scalar-identity structure, where all state dimensions within a single attention head share an identical decay dynamic \citep{daoTransformersAreSSMs2024}. While computationally efficient, this structural compromise inherently limits the independent representational capacity of the model's internal dimensions.
Despite their recurrent formulation, bounded-depth SSMs operate under strict theoretical ceilings. \citet{merrillIllusionStateStateSpace2025} point out that modern linear SSMs also reside in the $TC^0$ complexity class and cannot compute non-commutative permutation compositions. Consequently, their recurrent state can act as an ``illusion'' in complex tracking tasks \citep{merrillIllusionStateStateSpace2025}. The non-negative transition gates required for training stability mathematically prevent models from computing periodic or modulo properties \citep{sarrofExpressiveCapacityState2024}. Furthermore, \citet{jelassiRepeatMeTransformers2024} found that fixed-size latent state of SSMs restricts exact information copying from long contexts compared to Transformers.

However, structural dependencies in human language are practically bounded, with syntactic centre-embedding rarely exceeding three levels \citep{karlssonConstraintsMultipleCenterEmbedding2007}. Within these bounded contexts,\citet{sarrofExpressiveCapacityState2024} theoretically prove that a two-layer SSM can model bounded hierarchical structures without simulating an explicit Last-In-First-Out (LIFO) stack. Instead of maintaining a true stack, the architecture tracks nesting depth via an unbounded counter and relies on local state tracking to resolve matching elements.

\subsection{The Evaluation Gap: From State Tracking to Semantic Rollback}

Formal language theory provides a mathematical framework for isolating the specific memory mechanisms of neural architectures \cite{deletangNeuralNetworksChomsky2023a}. The Chomsky hierarchy \citet{chomskyThreeModelsDescription1956} categorises languages based on their required computational machinery. According to \citet{deletangNeuralNetworksChomsky2023a},regular languages demand only $O(1)$ memory, while context-free languages require a LIFO stack to manage hierarchical and nested dependencies. 

The standard Flip-Flop language evaluates pure semantic state maintenance across distance and is classified as a star-free regular language. The task requires a model to recall the boolean state of the most recent write instruction, demanding an $O(1)$ destructive overwrite memory mechanism that permanently replaces the preceding state. Transformers frequently fail on the Flip-Flop language due to continuous softmax attention dilution \citep{liuExposingAttentionGlitches2023}. In contrast, contemporary SSMs, as shown by \citet{sarrofExpressiveCapacityState2024} bypass heuristic attention approximations and can model the standard Flip-Flop language deterministically at finite precision.Despite this success, destructive overwrite mechanisms fail to capture the dynamic history maintenance required for natural communication. Discourse-level disfluencies and self-repair involve non-monotonic semantic updates where a speaker retracts a prior statement to restore a previous state. Strict left-to-right autoregressive generation struggles to accommodate these semantic revisions natively \citep{welleckNonMonotonicSequentialText}. Because standard monotonic benchmarks permanently annihilate prior states upon processing a new write instruction, they cannot evaluate a model's capacity for reversible historical recovery.

Conversely, the Dyck language serves as the standard benchmark for testing syntactic nesting and stack memory. Both Recurrent Neural Networks \citep{hewittRNNsCanGenerate2020} and self-attention networks \citep{yaoSelfAttentionNetworksCan2023} can process bounded hierarchical Dyck languages. However, the Dyck task exclusively evaluates structural well-nestedness, forcing the model to match closing brackets to opening ones while remaining entirely agnostic to the semantic payload of those states. Models do not dynamically extract or restore specific values during the structural nesting process.

This divergence between semantic state extraction in Flip-Flop tasks and structural management in Dyck languages creates a significant evaluative gap. Models trained without rigorous constraints tend to adopt superficial decision rules that succeed on in-distribution data but fail catastrophically under out-of-distribution conditions \cite{geirhosShortcutLearningDeep2020}. It remains unclear if the stack-free counting mechanism identified in two-layer SSMs \citet{sarrofExpressiveCapacityState2024} represents a robust algorithmic solution or a fragile statistical shortcut. Determining whether these models can acquire true state management requires a combined approach that evaluates reversible semantic updates against the pressure of out-of-distribution generalisation.

%% file: sections/3Mthd.tex
% !TeX root = ../neurips_2025.tex
\section{Methodology}

\subsection{Task Formulation: The UNDO Flip-Flop Language}

We introduce the UNDO Flip-Flop task to evaluate non-monotonic, reversible state updates. 
%motivated by discourse-level semantic repair in human communication. Semantic repair refers to instances where a speaker retracts a prior assertion and reinstates an earlier state (Schegloff, 2013) — for example, "Set the value to 1... wait, make it 0... actually, scratch that." 
The UNDO Flip-Flop task extends the standard instruction set with an [UNDO] operator and operates on an implicit bounded stack. All existing constraints of the standard Flip-Flop language are preserved: every sequence begins with a write command and ends with a read command. Rather than operating on a simple overwritable register, the task now operates on an implicit bounded stack. The operational semantics are as follows:

\begin{itemize}
    \item \textbf{write (Push)}: Pushes the boolean value $v$ onto the history stack. The active computational state becomes $v$.
    \item \textbf{ignore (No-op)}: A null operation that leaves the underlying stack and active state completely unchanged.
    \item \textbf{[UNDO] (Pop)}: Retracts the most recent write operation by removing the top element from the stack. The active state deterministically reverts to the previously stored historical value.
    \item \textbf{read (Peek)}: Evaluates the model by requiring it to explicitly output the current active state residing at the top of the stack.
\end{itemize}

%The inclusion of the [UNDO] operator causes a verifiable shift in the underlying computational complexity. Because the model must now retrieve specific historical states after processing an arbitrary sequence of intermediate writes and retractions, the task escapes the boundaries of regular languages. In an unbounded setting, UNDO Flip-Flop becomes a context-free language requiring the memory capacity of a pushdown automaton. The operational sequence maps algebraically to a Dyck-style bracket discipline, where a write command functions as an opening bracket and an [UNDO] command functions as a closing bracket.
The inclusion of the [UNDO] operator elevates the computational complexity beyond regular languages. The operational sequence maps algebraically to a Dyck-style bracket discipline, where a write command functions as an opening bracket and an [UNDO] command functions as a closing bracket. However, unlike classic Dyck tasks that test structural well-nestedness, UNDO Flip-Flop isolates the computational cost of semantic state retrieval under dynamic retractions. It serves as a controlled probe for reversibility.

\begin{table}[h]
\centering
\begin{tabular}{lccccccc}
\toprule
Step & 1 & 2 & 3 & 4 & 5 & 6 & 7 \\
\midrule
Operation 
& $W(1)$ & $W(0)$ & $I$ & $W(1)$ & $\texttt{UNDO}$ & $I$ & $R$ \\

Stack 
& $[1]$ & $[1,0]$ & $[1,0]$ & $[1,0,1]$ & $[1,0]$ & $[1,0]$ & \\

\bottomrule
\end{tabular}
\caption{Example of the \texttt{[UNDO]} Flip-Flop. Final output: $0$.}
\end{table}

\subsection{Data Generation and Syntactic Constraints}
The task data adheres to strict structural constraints. To prevent illegal empty-stack retractions, an \texttt{[UNDO]} token is strictly valid only when the active stack depth is greater than one. This condition ensures that the number of retractions never exceeds the number of active writes in any sequence prefix, directly satisfying the prefix-balance condition of Dyck grammars.

Following \citet{liuExposingAttentionGlitches2023}, the dataset maintains specific background sparsity settings. The probability of generating an ignore instruction is $p_i = 0.8$, and a read instruction is $p_r = 0.1$. The remaining probability mass is divided equally between writes ($p_w = 0.05$) and retractions ($p_u = 0.05$). This distribution ensures the \texttt{[UNDO]} operator appears frequently enough to be learnable without preventing the accumulation of meaningful history.Sequences are partitioned into two sets by length. In-distribution (ID) sequences contain 1 to 50 tokens for training and baseline evaluation. Out-of-distribution (OOD) sequences range from 51 to 100 tokens and are used exclusively to test length generalisation. The task is structured as a deterministic prediction challenge where training labels mask all positions except those immediately following a read token, using the ignore index \texttt{-100}. This isolates the loss computation, forcing the model to optimise strictly for semantic state recovery.

\subsection{Model Architecture and Training Regimen}
We evaluate the Mamba-2 state space model \citep{daoTransformersAreSSMs2024}. As discussed in Section~\ref{sec:models-formal-expressivity}, the scalar-identity structure of Mamba-2 trades per-dimension expressivity for computational efficiency. We proceed under the working hypothesis that Theorem 6's capacity result \citet{sarrofExpressiveCapacityState2024} extends to Mamba-2, noting that the scalar-identity simplification may impose additional constraints not captured by the theorem. Establishing this formally is left for future work. 

To isolate algorithmic capacity from statistical memorisation, we employ deliberately small, dimension-constrained configurations ($d_{\text{model}} = 16$, $d_{\text{state}} = 16$, $d_{\text{conv}} = 4$, expansion factor = 2). Following \citet{sarrofExpressiveCapacityState2024}'s theoretical framework, we compare 1-layer and 2-layer variants. This directly tests whether the structural simplifications of Mamba-2 impair the stack-free counting mechanism theoretically proven to exist in 2-layer SSMs (see Section~\ref{sec:models-formal-expressivity}).Models are trained using AdamW with a learning rate of $3 \times 10^{-4}$ and a batch size of 16. Training continues until full convergence on the in-distribution set, evaluated every 100 steps. This strict convergence criterion ensures that any observed generalisation failure stems from algorithmic limits rather than under-training.

\subsection{Evaluation Metrics and Mechanistic Probing Design}
\label{subsec: probe}
The primary evaluation metric is sequence-level exact match accuracy. A sequence is marked correct strictly if the model accurately predicts the required boolean state at every read position. This prevents partial successes from masking systematic failures in state tracking.To isolate non-monotonic state management from generic length generalisation limits, we also design the Aggressive UNDO Pressure Test. This mechanistic probe maximises structural complexity while holding sequence length fixed at 50 tokens, being within the training distribution. 

Probe sequences are constructed in three phases: 

\begin{enumerate}
    \item \textbf{Build Phase:} $D+1$ consecutive writes populate the implicit stack to depth $D+1$.
    \item \textbf{Undo Phase: } $D$ consecutive \texttt{[UNDO]} commands execute a deep, continuous rollback without intervening writes.
    \item \textbf{Tail Phase:}  Remaining tokens are filled with ignore instructions, terminating in a final read.

\end{enumerate}

Our probe fixes the rollback depth at $D=10$. This requires a minimum sequence length of $33$. By maintaining total sequence lengths under 50, any observed accuracy collapse provides direct evidence of algorithmic failure independent of length generalisation.

%Finally, we track two specific error signatures. \textbf{The Deep History Loss Rate} measures the frequency at which the model fails to retrieve a historical state buried more than one step deep. Concurrently, the \textbf{Local Toggle Heuristic Rate} tracks instances where the model incorrectly predicts the outcome by simply applying a boolean inversion to the most recently observed bit, circumventing true historical retrieval.

Finally, we use two specific causal probes to identify the underlying computational strategies. The Deep History Loss Rate is designed to detect whether the model successfully maintains access to historical states buried beneath the current stack top. Conversely, the Local Toggle Heuristic Rate identifies reliance on the most recently observed bit by measuring the model's sensitivity to perturbations of discarded (popped) values. By applying these probes to sequences where the model achieves correct outputs, we can distinguish between genuine state management and superficial shortcuts that happen to align with the task logic.

%% file: sections/4Experiments.tex
% !TeX root = ../neurips_2025.tex
\section{Experiments}

\begin{table}[t]
\centering
\small
\caption{Performance on standard and \texttt{[UNDO]} Flip-Flop tasks.}
\label{tab:flipflop_experiments}
\begin{tabularx}{\linewidth}{@{}>{\raggedright\arraybackslash}X>{\centering\arraybackslash}p{0.11\linewidth}>{\centering\arraybackslash}p{0.14\linewidth}>{\centering\arraybackslash}p{0.17\linewidth}>{\centering\arraybackslash}p{0.17\linewidth}@{}}
\textbf{Model Variant} & \textbf{Epoch} & \textbf{Train Acc. (\%)} & \textbf{ID Test Acc. (\%)} & \textbf{OOD Test Acc. (\%)} \\
\midrule
Baseline standard Flip-Flop, One-Layer & 25 & 100 & 100 & 96.25 \\
Baseline standard Flip-Flop, Two-Layer & 5 & 100 & 100 & 99.05 \\
\texttt{[UNDO]} Flip-Flop, One-Layer & 100 (max) & 95.95 & 95.95 & 79.10 \\
\texttt{[UNDO]} Flip-Flop, Two-Layer & 100 (max) & 98.55 & 98.55 & 87.35 \\
\end{tabularx}
\end{table}

\subsection{Baseline Performance on Monotonic State Tracking}

The standard Flip-Flop task establishes a computational baseline for monotonic, destructive overwrite memory. The star-free case for which \citet{sarrofExpressiveCapacityState2024} prove that SSMs possess length-generalising solutions (Theorems 1 and 4). Both model variants converged without difficulty. The 1-layer Mamba 2 model reached 100\% in-distribution accuracy within 25 epochs, maintaining 96.25\% on out-of-distribution sequences. The 2-layer model converged within 5 epochs and achieved 99.05\% out-of-distribution accuracy. These results are broadly consistent with \citet{sarrofExpressiveCapacityState2024} empirical findings for Mamba 1 on the same task, suggesting that Mamba 2's scalar-identity architectural change does not measurably impair performance on star-free state tracking. While this does not rule out architecture-specific confounds on harder tasks, it establishes that the model is capable of basic state tracking under these training conditions, providing a methodological baseline.

\begin{figure}[htbp]
    \centering
    \includegraphics[width=0.8\textwidth]{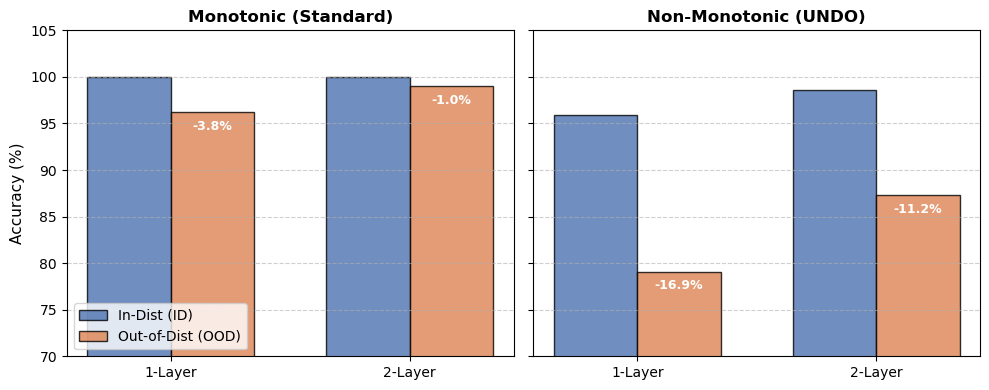}
    \caption{Asymmetric generalisation failure under UNDO}
    \label{fig:std_vs_undo}
    
\end{figure}

\subsection{Performance Degradation on Reversible State Updates}
As established in Section~\ref{sec:rl}, the operational structure of \texttt{[UNDO]} Flip-Flop maps onto a bounded Dyck discipline. While \citet{sarrofExpressiveCapacityState2024}'s Theorem 6 suggests that, under the general SSM framework, a two-layer SSM is theoretically able to process such hierarchical structures, \texttt{[UNDO]} Flip-Flop imposes the additional computational demand of continuous semantic retrieval at every read position. Our results indicate that models struggle to discover this exact mechanism under gradient descent. As Table \ref{tab:flipflop_experiments} shows, 2-layer model achieved an in-distribution accuracy of 98.55\%, outperforming the 1-layer model's 95.95\%. This performance gap aligns with the theoretical prediction that additional depth confers structural capacity. However, out-of-distribution performance dropped sharply to 87.35\% for the 2-layer model and 79.10\% for the 1-layer variant. The 2-layer model's OOD decline of 11.2 percentage points contrasts starkly with the less than one percentage point drop observed on the standard Flip-Flop task, marking a qualitative shift in generalisation failure.

This degradation is unlikely to reflect insufficient training. While standard Flip-Flop variants converged rapidly, both \texttt{[UNDO]} Flip-Flop models failed in reaching perfect accuracy after the full 100-epoch training limit. This struggle suggests the architectures reached the optimisation limits of gradient descent for this specific task. Furthermore, the near-perfect generalisation on the monotonic baseline rules out generic length generalisation difficulty. The 11–17 percentage point excess OOD decline is specifically attributable to the non-monotonic computational demands of the \texttt{[UNDO]} operator.

\subsection{Mechanistic Interpretation}

\begin{figure}[htbp]
\centering
\begin{tikzpicture}[node distance=1.2cm, >=stealth, font=\small]

% --- Left Side: Ground Truth (Stack) ---
\node[font=\bfseries] at (0, 0.8) {Ideal: Stack-based Pop};
\draw[thick, ->] (0,0.5) -- (0,-5.5);

\node (s1) at (0,0) [draw, rectangle, minimum width=2.5cm] {W(0) $\rightarrow$ \textbf{State: 0}};
\node (s2) at (0,-1.2) [draw, rectangle, minimum width=2.5cm] {W(1) $\rightarrow$ \textbf{State: 1}};
\node (s3) at (0,-2.4) [draw, rectangle, minimum width=2.5cm, fill=blue!10] {W(1) $\rightarrow$ \textbf{State: 1}};
\node (s4) at (0,-3.6) [draw, rectangle, minimum width=2.5cm, fill=green!10] {UNDO $\rightarrow$ \textbf{State: 1}};
\node (s5) at (0,-4.8) [draw, rectangle, minimum width=2.5cm, fill=green!10] {UNDO $\rightarrow$ \textbf{State: 0}};

\node[left=0.5cm of s3, font=\scriptsize, align=right] {Stack:\\ \texttt{[0, 1, 1]}};
\node[left=0.5cm of s4, font=\scriptsize, align=right] {Pop 1 $\rightarrow$\\ \texttt{[0, 1]}};
\node[left=0.5cm of s5, font=\scriptsize, align=right] {Pop 1 $\rightarrow$\\ \texttt{[0]}};

% --- Right Side: Shortcut (Toggle) ---
\begin{scope}[xshift=5.5cm]
\node[font=\bfseries] at (0, 0.8) {Learned: Local Toggle Heuristic};
\draw[thick, ->] (0,0.5) -- (0,-5.5);

\node (e1) at (0,0) [draw, rectangle, minimum width=2.5cm] {W(0) $\rightarrow$ \textbf{State: 0}};
\node (e2) at (0,-1.2) [draw, rectangle, minimum width=2.5cm] {W(1) $\rightarrow$ \textbf{State: 1}};
\node (e3) at (0,-2.4) [draw, rectangle, minimum width=2.5cm, fill=blue!10] {W(1) $\rightarrow$ \textbf{State: 1}};
\node (e4) at (0,-3.6) [draw, rectangle, minimum width=2.5cm, fill=red!20, draw=red, thick] {UNDO $\rightarrow$ \textbf{State: 0}};
\node (e5) at (0,-4.8) [draw, rectangle, minimum width=2.5cm, fill=red!20, draw=red, thick] {UNDO $\rightarrow$ \textbf{State: 1}};

\node[right=0.5cm of e4, color=red, font=\scriptsize, align=left] {\textbf{Divergence!}\\ Inverts $1 \rightarrow 0$};
\node[right=0.5cm of e5, color=red, font=\scriptsize, align=left] {Inverts $0 \rightarrow 1$};
\end{scope}

% Connecting line for divergence
\draw[dashed, red, thick] (s4.east) -- (e4.west) node[midway, above, font=\tiny] {Mismatch};

\end{tikzpicture}
\caption{Mechanistic comparison between the robust stack mechanism (left) and the observed toggle heuristic (right). The shortcut fails to resolve consecutive identical bits because it treats [UNDO] as a bitwise inversion of the current state rather than a historical retrieval.}
\label{fig:toggle_heuristic}
\end{figure}
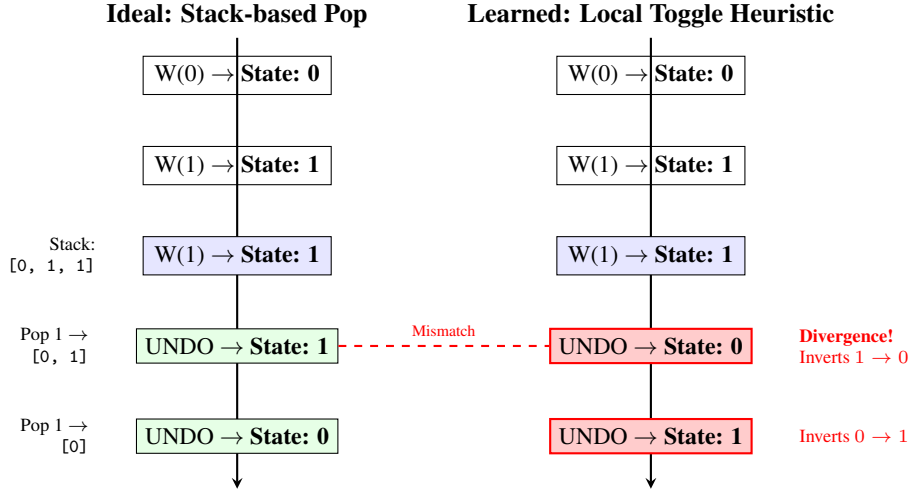

We applied the Aggressive UNDO Pressure Test to the 2-layer model to scrutinise the nature of its internal representations. Under this dense retraction pressure, sequence-level accuracy collapsed to $41.10\%$. As the task requires a binary prediction, this performance falls significantly below the $50\%$ accuracy expected from random chance. Such a result indicates that the model is not merely guessing but is actively and systematically producing incorrect outputs. As demonstrated in studies of syntactic heuristics, models frequently rely on superficial strategies that fail to reflect the intended underlying logic, often leading to systematic errors on adversarial ``challenge sets'' \citet{mccoyRightWrongReasons2019a}. To investigate this possibility, we conducted a causal ablation study on the 552 sequences where the model produced entirely correct predictions.

The results provide evidence of ``shortcut learning'' \citep{geirhosShortcutLearningDeep2020}. Among the correctly predicted samples, the Local \textbf{Toggle Heuristic Rate} reached $38.04\%$, whereas the \textbf{Deep History Loss Rate} was only $2.17\%$. This discrepancy indicates that even when the output matches the ground truth, the model often relies on inverting the most recently observed bit rather than performing genuine historical retrieval. The marginal history loss rate confirms that the architectural capacity for signal retention is not the primary bottleneck. Instead, the optimisation process has favoured a locally viable toggle heuristic that fails to specialise when consecutive identical bits or deep rollbacks are encountered.

\begin{figure}[htbp]
    \centering
    \includegraphics[width=0.8\textwidth]{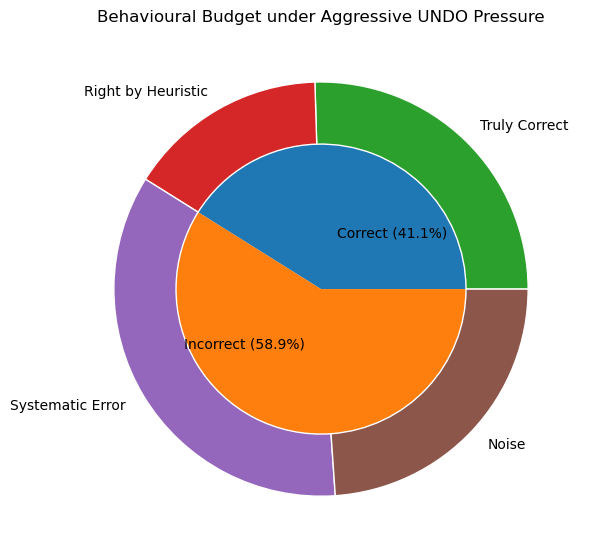}
    \caption{Behavioural Decomposition Chart}
    \label{fig:example_image}
    
\end{figure}

These findings suggest that the internal state of the Mamba-2 model acts as a fragile proxy for a stack. Although as \citet{sarrofExpressiveCapacityState2024} prove that the two-layer architecture possesses the theoretical capacity to model bounded hierarchical structures, gradient-based optimisation appears to converge on simpler bit-flipping strategies. This heuristic is statistically sufficient for the sparse training distribution but lacks the algorithmic depth required for non-monotonic semantic recovery. The observed performance in the pressure test is therefore a result of accidental alignment between a shortcut and the task logic, rather than a manifestation of true state management.

%% file: sections/5Discussion.tex
% !TeX root = ../neurips_2025.tex
\section{Discussion}

The findings point out that gradient descent does not reliably find the correct \texttt{UNDO} solution, instead of Mamba-2 cannot represent it.  The low Deep History Loss Rate rules out storage as the bottleneck: the failure occurs at retrieval. This places the finding in a different category from the theoretical ceiling described by \citet{merrillIllusionStateStateSpace2025}, whose $TC^0$  analysis identifies what linear SSMs cannot express at all. The present work concerns what optimisation reliably extracts from what is, in principle, expressible — a failure mode that theoretical expressivity results alone cannot predict or detect.

The toggle heuristic likely emerges from the interaction of two factors. The first is distributional: the sparse retraction probability, combined with the dominance of ignore tokens, means that isolated, shallow UNDO operations are overwhelmingly typical during training. In such cases, inverting the current bit and correctly popping the stack produce the same output, so the model receives consistent gradient signal for the heuristic without ever being forced to distinguish it from genuine historical retrieval. The second factor is structural. Mamba-2's scalar-identity constraint biases the architecture towards linear operations on a single scalar quantity, which a sign-flip satisfies directly; an ordered, addressable stack of historical values does not. This structural argument remains speculative, however, since the extension of Theorem 6 to Mamba-2 is a working hypothesis. If the scalar-identity constraint makes the correct solution inexpressible altogether, the failure is one of expressibility rather than learnability, and the two diagnoses carry substantially different implications.

Several limitations bound what can be concluded. The evaluation covers a single Mamba-2 configuration, and the SSM lineage has since advanced to Mamba-3, whose architectural revisions may alter the balance between expressibility and optimisation dynamics that are not addressed here. All experiments were conducted on a single random seed, so the reported accuracy figures carry no variance estimates; given that the model appears to occupy a suboptimal basin on the \texttt{UNDO} task, performance may vary meaningfully across initialisations. The limited computational resource is another factor. We also admit the lack of a Transformer baseline on the \texttt{UNDO} Flip-Flop task, which would have clarified whether the toggle heuristic is particular to SSM dynamics or a more general failure of gradient descent on this problem class. Finally, the training distribution is itself a design choice: the sparse retraction probability that allowed the toggle heuristic to become viable was also what kept the task learnable at all under the given configuration. A denser schedule of consecutive retractions might suppress the shortcut but could equally destabilise training under the same architecture.

The implications of these findings extend to a range of natural language tasks that require non-monotonic semantic updates. The most direct case is speech disfluency and repair, both within a single utterance (a speaker retracts a constituent mid-production and restates it) and across turns (a prior commitment is explicitly withdrawn and a previous communicative state must be restored). A model relying on toggle logic would interpret retraction as semantic inversion rather than as recovery of a prior state, producing a systematic rather than probabilistic error in dialogue understanding. Beyond repair, the failure mode is relevant to any reasoning task where inference is not strictly cumulative. Chain-of-thought processes that require abandoning an incorrect intermediate conclusion and recovering a prior position place the same demand on ordered historical access that the toggle heuristic cannot meet. Counterfactual and conditional reasoning, in which a model must suppress a conclusion that follows from a premise now being retracted, is structurally equivalent to a deep rollback under the formalism developed here. If SSMs are predisposed to shortcut this operation through local inversion, their reliability in extended multi-step reasoning tasks deserves closer scrutiny, independent of their strong performance on standard sequential benchmarks.

%% file: sections/6Conclude.tex
% !TeX root = ../neurips_2025.tex
\section{Conclusion and Future Work}

This paper introduced the UNDO Flip-Flop task as a controlled probe for reversible semantic state management and evaluated Mamba-2 under this framework. The results show that both model variants fail to acquire the stack-based rollback mechanism that \citet{sarrofExpressiveCapacityState2024}'s Theorem 6 proves to be expressible, converging instead on a local toggle heuristic that inverts the current bit rather than retrieving a stored historical value. The primary contribution is a distinction between architectural expressibility and what gradient-based optimisation reliably learns: the low Deep History Loss Rate confirms the former is not the bottleneck, while the systematic below-chance failure under adversarial rollback pressure confirms the latter is.

Three directions follow directly from the limitations of this work. The most pressing is a formal analysis of whether Theorem 6 extends to Mamba-2's scalar-identity architecture, since the expressibility question remains an unverified working hypothesis that determines how the empirical failure should be interpreted. On the experimental side, replication across multiple random seeds, a Transformer baseline, and evaluation on Mamba-3 would establish how robust and architecture-specific the observed shortcut is. Finally, it remains an open question whether modifying the training distribution, for instance, through a curriculum that progressively increases retraction density, is sufficient to suppress the toggle heuristic and force the model towards a more robust solution.

%% file: sections/7Appendex.tex
% !TeX root = ../neurips_2025.tex
\section{Appendix}

\begin{figure}[htbp]
    \centering
    \includegraphics[width=0.9\textwidth]{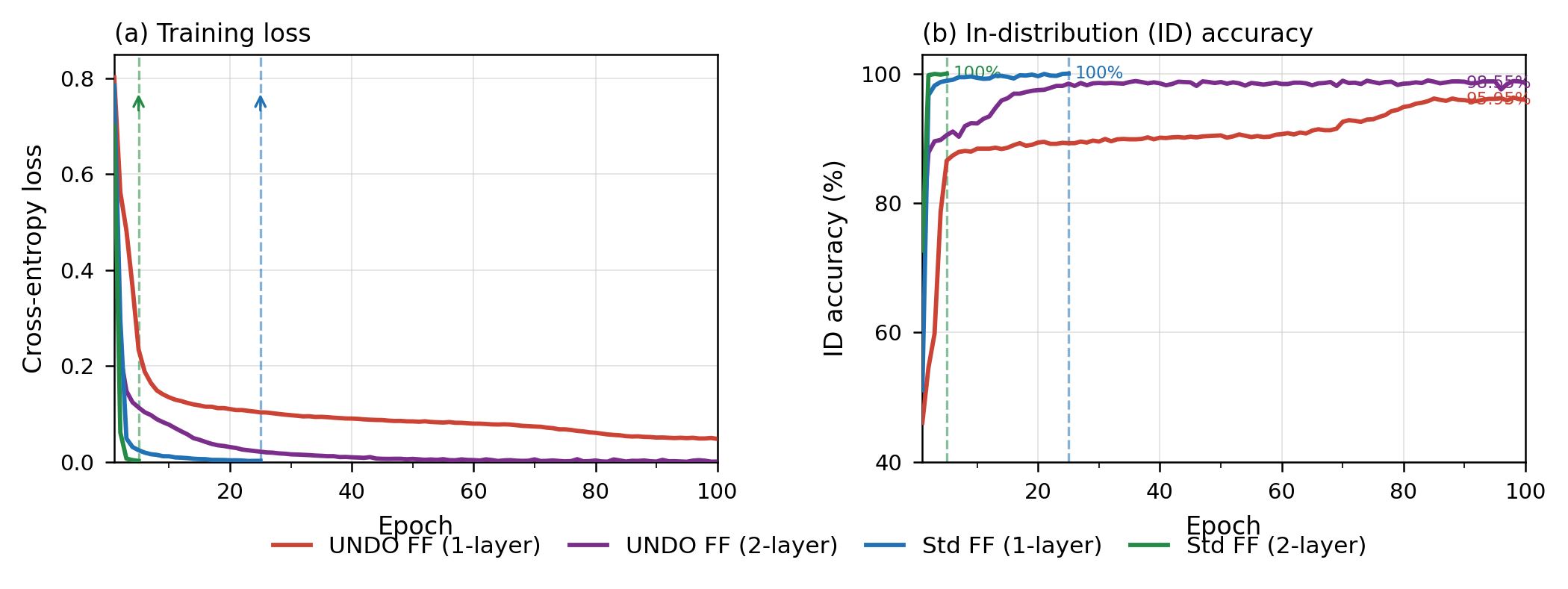}
    \caption{Training Convergence}
    \label{fig:convergence}
    
\end{figure}